# Feature Extraction and Soft Computing Methods for Aerospace Structure Defect Classification

Gianni D'Angelo, Salvatore Rampone
University of Sannio
Dept. of Science and Technology
Benevento, Italy
{dangelo, rampone}@unisannio.it

___________________________________________________________________

*Abstract*

This study concerns the effectiveness of several techniques and methods of signals processing and data interpretation for the diagnosis of aerospace structure defects. This is done by applying different known feature extraction methods, in addition to a new CBIR-based one; and some soft computing techniques including a recent HPC parallel implementation of the U-BRAIN learning algorithm on Non Destructive Testing data. The performance of the resulting detection systems are measured in terms of Accuracy, Sensitivity, Specificity, and Precision. Their effectiveness is evaluated by the Matthews correlation, the Area Under Curve (AUC), and the F-Measure. Several experiments are performed on a standard dataset of eddy current signal samples for aircraft structures. Our experimental results evidence that the key to a successful defect classifier is the feature extraction method - namely the novel CBIR-based one outperforms all the competitors – and they illustrate the greater effectiveness of the U-BRAIN algorithm and the MLP neural network among the soft computing methods in this kind of application.

___________________________________________________________________

## I. INTRODUCTION

The use of composite materials, particularly carbon fiber reinforced polymer (CFRP), in the aerospace industry is growing rapidly, especially in the production of the components subjected to heavy loads and efforts. Due to their unique mechanical properties, namely, high strength-to-weight ratio, high fracture toughness, and excellent corrosion resistance properties, they are used at critical points in the construction of an aircraft [1, 2]. They are widely used in the outer covering of the aircraft, such as flaps, hatches, sides of the engine, floors, rudders, elevators, ailerons etc. The composite material design and manufacturing technologies have matured to a level that Boeing Company is using composite material for 50% of the primary structure in its 787 program. There is also a growing interest in carbon-fiber reinforced aluminum (FRA), which is stronger than aluminum and cheaper and lighter than steel. For example these materials are developed as part of the Future Advanced Rotorcraft Drive System (FARDS) program as a direct replacement for the existing steel liners that are commonly used today in rotorcraft transmissions [3]. Unfortunately, there is a great variety of possible manufacturing defects that regards these materials [4]. The most widespread types of defects are the following:

- Delamination between plies of outer skin, parallel to surface;
- Matrix crack;
- Disbanding between the outer skin and the honeycomb core;
- Fiber fracture;
- Cracked honeycomb core parallel to the inspection surface;
- Crushed honeycomb core in parallel to the area;
- Disbanding between inner skin and honeycomb core;
- Fluid ingress in honeycomb core.
- Damages induced by the stress, environment influences and others.
- Wear, scratch, indentation and cleft
- Creep deformation.

These defects are difficult to diagnose and the analysis is strongly influenced by many factors that may also arise from the complexity of manufacturing processes. In addition, some techniques of inspection and/or some detection equipment may have systematic errors or accidental ones. Most of the maintenance processes are conducted by human inspectors, whose individual experiences may yield to differences in result interpretation. Reliable human performance is crucial to inspections and tests. Inadequate human performance could lead to missed defects and inaccurate reports, with potentially serious safety and cost consequences. In addition, manpower assigned to such tasks results in significant recurrent costs and it is time consuming. For this reason, the accuracy of diagnosis of aerospace materials is better entrusted to objective testing and advanced data interpretation methods.

Non-Destructive Testing (NDT) allows one to implement a control over the material at different stages of its evolution and permits to safeguard the integrity of the structure during the analysis. Visual and strike method, optical holography, X-ray, ultrasonic wave, eddy current testing and infrared detection, X-ray and ultrasonic C-scan are the most common methods. Due to the heterogeneity of the composite structure, the NDT of composites is very complex and sometimes several methods will take to test the same component [5]. The analysis of the set of signal informative parameters, i.e. performing multi-parameter control, is one of the possible ways to increase effectiveness and reliability of the non-destructive testing of composites. Furthermore, the accuracy of diagnosis of composite materials is determined not only by the physical methods used to obtain experimental data, but also by the data processing methods [6]. Spectrum analysis and pattern recognition are often used in multi-parameter control [7]. However, the application of these methods requires sophisticated techniques for processing signals that lead to the solution of nonlinear equations complex with a high number of variables [8]. The difficult and sometimes impossible solution for these equations leads to a reduction in the efficiency of the system of NDT. These difficulties also do not allow the automation of the test and deprive them of the same dynamism typical of a system able to adapt to changes in the parameters of the testing system at run-time. NDT of composites should be performed with methods able to collect the most comprehensive information about new defects, expand existed base of defects and increase diagnostics system precision at runtime. Finally, the processing has to deal with a great amount of data when multiple elements are processed at the same time.

An alternative method of data processing and construction of decision rules for multi-parameter NDT of composite materials is to use Soft Computing techniques [9]. Soft computing is the combination of methodologies intended to model and make possible solutions to real world challenging problems, which are not modeled or too complex for mathematical modeling. Its aspiration is to utilize the tolerance for approximation (model features are similar to the real ones but not the same), uncertainty (not sure that the model features belief are the same as that of the entity), imprecision (model features quantities are not same as real ones but close to them) and partial truth in order to achieve close resemblance with human-like decision making. The guiding theory of soft computing is to use this tolerance to achieve, robustness tractability and low solution cost. Human mind is the role model for soft computing. Some of the soft computing techniques are Artificial Neural Network (ANN), Fuzzy Logic (FL), Adaptive Neuro-Fuzzy Inference System (ANFIS), and evolutionary computation [9]. Soft Computing methods as ANN are proven to be effective in non-destructive testing [10]. In recent years Support vector machines (SVMs) showed comparable or better results than ANNs and other statistical models [11], and they are mostly used to classify the defects [12]. In [13] the authors proposed a method based on the spectrum analysis and on a proper algorithm that uses a soft computing technique.

However, as we will evidence in this work the key to a successful soft-computing based testing system is to choose the right feature extraction method representing the defect as accurately and uniquely as possible in a short time.

This study concerns the effectiveness of several techniques and methods of signals processing and data interpretation for the diagnosis of aerospace structure defects. This is done by applying different known feature extraction methods, and a novel CBIR-based one; and some soft computing techniques including a recent HPC parallel implementation of the U-BRAIN learning algorithm to NDT data. The performance of the resulting detection systems are measured in terms of Accuracy, Sensitivity, Specificity, and Precision. Their effectiveness are evaluated by the Matthews correlation, the Area Under Curve (AUC), and the F-Measure. Several experiments are performed on a standard dataset of Eddy Current (EC) signal samples for aircraft structures. Eddy current testing is one of the most extensively used non-destructive techniques for electrically inspecting materials at very high speeds that does not require any contact between the test piece and the sensor [14].

The paper is organized as follows. Section II is an overview on the soft computing techniques employed. In Section III we describe the eddy current testing as a NDT case study and the way to characterize the defects. In Section IV the feature extraction methods involved are described. In Section V the experimental method and the data set used are outlined. Experimental results are shown in Section VI. Conclusions are drawn in Section VII.

II. SOFT COMPUTING BASED DATA PROCESSING

The main challenges in handling NDT results lie in finding a correspondence between the measured data and a specifically kind of defect. To this aim several data analysis techniques have been traditionally used, including regression analysis, cluster analysis, numerical taxonomy, multidimensional analysis, multivariate statistical methods, stochastic models, time series analysis, nonlinear estimation techniques, and others [15]. Unfortunately many of these techniques have inherent limitations. For example, a statistical analysis can determine correlations between variables in data, but cannot evidence a justification of these relationships in the form of higher-level logic-style descriptions and laws. To overcome the above limitations, researchers have turned to ideas and methods developed in Machine Learning [16], whose goal is to develop computational models for acquiring knowledge starting from facts and background knowledge. These and related efforts have led to the emergence of a new research area, frequently called Data Mining (DM) and Knowledge Discovery in Databases (KDD) [17, 18]. In the Machine Learning approach, an algorithm - usually off line - 'learns' about a phenomenon by looking at a set of occurrences (used as examples) of that phenomenon. Based on these, a model is built and can be used – on line - to predict characteristics of future (unseen) examples of the phenomenon. The whole operating scenario is depicted in Fig. 1 where all the off-line activities, associated to the classification frameworks, are reported into a grey box, whereas the other ones can be managed on-line within the context of a real-time detection system. However, in order to keep the classifier up-to-date with the newest data, periodical re-training is required.

Specifically we use Machine Learning techniques falling in the Soft Computing area [9]. In this way they are tolerant of imprecision, uncertainty, partial truth, and approximation. In this study we apply several Soft Computing tools including rule-based methods (C4.5/J48 [19]), ANNs (MultiLayer Perceptron (MLP) [20]), Bayesian networks (Naive Bayes classifier [21]), and Learning Algorithms (Uncertainty managing Batch Relevance based Artificial Intelligence algorithm (U-BRAIN) [22]).

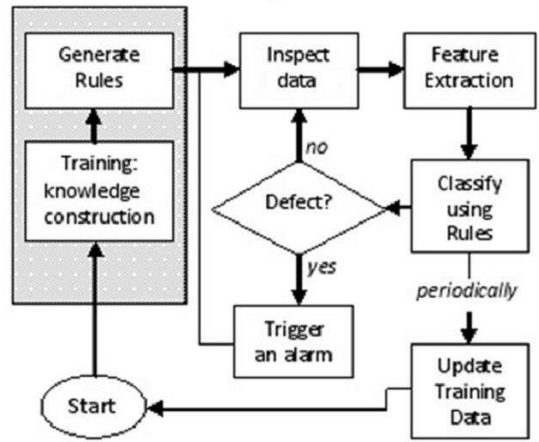

Fig. 1. Detection system architecture.

*A. NBC (Naïve Bayes Classifier)*

The NBC is a simple probabilistic classifier. Parameters used in the Naive Bayes model are determined from the training set using maximum likelihood algorithm. This model is then used along with a maximum a posteriori decision rule [23].

*B. MLP (Multilayer Perceptron)*

ANNs are mathematical models that simulate the structural/functional aspect of biological neural networks. A Multi Layer Perceptron (MLP) is a feed-forward ANN that consists of multiple layers of processing elements (nodes) in a directed graph, where each layer is fully connected to the next one. It is used for modeling complex relationship between input and output. MLP utilizes a supervised learning technique called back-propagation for training the network.

*C. C4.5/J48 algorithm*

The C4.5 algorithm builds tree structures from the training data. The rules extracted from the built tree are used to predict the class of the test data. One point of strength for the Decision Tree-based algorithms is that they can work well with huge data sets. We used the J48 open source Java implementation of the C4.5 algorithm in the Weka data mining tool.

*D. U-BRAIN (Uncertainty managing Batch Relevance based Artificial Intelligence algorithm)*

The U-BRAIN algorithm is a learning algorithm able to infer explicitly the laws that govern a process starting from a limited number of features of interest from examples, data structures or sensors. Each inferred rule is described as a Boolean formula (f) in Disjunctive Normal Form (DNF) [24], of approximately minimum complexity, that is consistent with a set of data. Such formula can be used to forecast the future process behavior. In its latest version, U-BRAIN can also act on incomplete data. Recently a parallel implementation of the algorithm has been developed by a Single Program Multiple Data (SPMD) [25] technique together to a Message-Passing Programming paradigm [26]. Algorithm details are reported in the Appendix I.

### III. CASE STUDY: EDDY CURRENT INSPECTION AND DEFECT CHARACTERIZATION

In aircraft manufacturing and maintenance, Eddy Current inspection [14] is one of several NDT methods widely used for evaluating the property of materials, components, systems, without causing damage during the analysis. EC inspection uses the electromagnetism principle as the basis for conducting examinations. EC inspection appears particularly suitable for FRA materials. Eddy currents are created through the process of electromagnetic induction. In an eddy current probe, an alternating current flows through a wire coil and generates an oscillating magnetic field. If the probe and its magnetic field are brought close to a conductive material like a metal test piece, a circular flow of electrons, known as an eddy current, will begin to move through the metal like swirling water in a stream. That eddy current flowing through the metal will in turn generate its own magnetic field, which will interact with the coil and its field through mutual inductance. Changes in metal thickness or defects like near-surface cracking will interrupt or alter the amplitude and pattern of the eddy current and the resulting magnetic field. This in turn affects the movement of electrons in the coil by varying the electrical impedance of the coil. Let's note that the presence of defects in a material in the most of interesting cases leads to a significant alteration of its electrical characteristics. So, changing material parameters corresponds to a particular output signal that is characterized by a specific frequency spectrum. The presence of damage is characterized by the changes in the signature of the resultant output signal that propagates through the structure and then in the probe coil.

One of the major advantages of EC as an NDT tool is the variety of inspections and measurements that can be performed. ECs can be used for crack detection, material thickness measurements, coating thickness measurements, conductivity measurements, material identification, heat damage detection, and damage depth determination. Furthermore EC testing is sensitive to small cracks, the inspection gives immediate results, the equipment is very portable and this method can be used for much more than flaw detection. In addition the test probe does not need to contact the part and is able to inspect complex shapes and sizes of materials. Nevertheless, a visual interpretation is generally used to analyze the data. Then, the results are influenced by subjectivity of human personnel. A more accurate data analysis can be obtained by solving complex multi-parametric partial differential equations. So, defect classification is generally carried out by signatures of the signal in the impedance plane, in the Fourier transform [27] or in Wavelet-based Principal Component Analysis (PCA) [28].

Here in order to characterize a defect, the output signal is firstly pre-processed by a feature extraction process, and then the extracted features are used as input to soft computing based classifiers.

IV. FEATURE EXTRACTION

Feature Extraction is a general term for methods of deriving values (features) intended to be informative, from an initial set of measured data. The set of extracted features is called Feature Vector. Feature extraction is related to dimensionality reduction [29].

This section contains brief descriptions of the pre-processing methods that were employed in this work as feature extraction strategies for EC signals, i.e. Fourier transform, Principal Component Analysis, Linear Discriminant Analysis, Wavelet transform, and Content Based Image Retrieval.

Principal Component Analysis and Linear Discriminant Analysis were applied in order to reduce the "curse of dimensionality" [30] effect.

Most of the information in a signal is carried by its transient phenomena and its irregular structures. In such cases it is preferable to decompose the signal into elementary building blocks that are well localized in both time and frequency. This alternative can be achieved by using the Short Time Fourier transform (STFT) [31] and the Wavelet Transform (WT) [32].

Content Based Image Retrieval (CBIR) aims to find invariances in images related to the same class of signals as class signatures [33].

*A. Fast Fourier Transform (FFT)*

One of the most common methods to analyze the frequency domain representation of a signal is the Fast Fourier Transform (FFT). Specifications about aerospace structure defects can be determined by examining the frequency spectrum of EC signals [34]. Mathematically FFT is the same as Discrete Fourier Transform (DFT), defined by:

$$X(e^{j\omega}) = \sum_{n=0}^{N-1} x(n) e^{-j\frac{2\pi}{N}kn} \quad k=0,....,N\text{-}1 \quad (1)$$

In equation (1), *x(n)* is the sampled version of collected data and *N* should be a power of two which is determined by the closest number to the window size. In this study, *N* is chosen to be 4096.

*B. Principal Component Analysis(PCA)*

Principal Component Analysis (PCA) is widely used in feature extraction to reduce the dimensionality of the raw data to a low-dimensional orthogonal features, while preserving information about prominent features and conserving the correlation structure between the process variables. PCA has found application in many fields such as face recognition [35], speech recognition [36], electroencephalogram signal classification [37] and, among others, NDT. It is a common technique for finding patterns in high volume data. PCA extracts orthogonal dominant features (Principal Components, PC) from a set of multivariate data. The dominant features retain most of the information by keeping the maximum variance of the features and the minimum reconstruction error. Each dominant feature is referred to as a vector of a eigenvectors space. Eigenvalues are scalar representations of the degree of variance within the corresponding PCs. PCs are ranked by their corresponding eigenvalues, and thus, the first PC captures the most significant variance in the dataset. The second PC is perpendicular to the first PC and it contains the next significant variance. In this work we use the eigenvectors as features. They are determined using the following steps [38]:

a) *subtraction of the mean*: the mean of the data is first subtracted from each of the data dimensions to produce a data set with zero mean. Then, the covariance matrix is calculated.

For *M* observations and *N* variables we have that the average is defined as:

$$\overline{X} = \frac{1}{M}\sum_{n=1}^{M} X_n \quad (2)$$

where $X_n$ is the $N$ dimensional column vector of the $n$-th observation.

b) *Covariance matrix calculation*. This is done by:

$$C = \frac{1}{M}\sum_{n=1}^{M}[(X_n - \bar{X}) \cdot (X_n - \bar{X})^T] = \frac{1}{M} A \cdot A^T \qquad (3)$$

where $A = [(X_1 - \bar{X}), (X_2 - \bar{X}), \dots (X_n - \bar{X})]$

Since the data is $N$ dimensional, the covariance matrix will be *NxN*.

c) *Eigenvectors extraction from covariance matrix*: since the covariance matrix is square, the covariance matrix is decomposed to obtain a matrix of eigenvectors which consists in the set of PCs. However, for large $N$, the determining of $N$ eigenvectors is an intractable task. So, a computationally feasible method to find these eigenvectors is generally adopted [39]. It consists in calculating the eigenvectors ($v_i$) of $A^T A$, indeed of $AA^T$, and retrieval the eigenvectors ($u_i$) of $C$ by:

$$u_i = A v_i \qquad (4)$$

These M eigenvectors are referred to as *eigensignals*. So, any signal can be identified as a linear combination of the eigensignals.

d) *Feature selection*: once eigenvectors are found from the covariance matrix, the next step is to order them by eigenvalue, from highest to lowest. This provides the components in order of significance. So, it is possible to ignore the components of lesser significance and the final data set will have less dimensions than the original.

e) *Deriving the new data set*: finally, the original feature space is multiplied by the obtained transition matrix (projection matrix), which yields a lower data dimensional representation.

C. *Linear Discriminant Analysis (LDA)*

Although PCA has a number of advantages, there are also some drawbacks [40]. One of them is that PCA gives high weights to features with higher variability disregarding whether they are useful for classification or not. Linear Discriminant Analysis (LDA) [29], on the other hand, searches for a dimensionally reduced vectors space while preserving as much of the class discriminatory information as possible. LDA takes into consideration the scatter of the data on both within-classes and between-classes. For all the samples of all classes two matrix are defined: one is called within-class scatter matrix, as given by :

$$S_w = \sum_{i=1}^{C} \sum_{j=1}^{N_i} (X_j - M_i)(X_j - M_i)^T \qquad (5)$$

where *C* is the number of classes, $M_i$ is the mean vector of the class *i*, $X_j$ is the *j*-th sample vector belonging to the class *i*, and $N_i$ is the number of samples in the class *i*.

The other matrix, called between-class scatter, is defined by:

$$S_b = \sum_{i=1}^{C} (M_i - M)(M_i - M)^T \qquad (6)$$

where M is the mean of all classes ($M=1/C \sum_i M_i$).

LDA computes a transformation that maximizes the between-class scatter while minimizing the within-class scatter. For a scatter matrix, the measure of spread is the matrix determinant. So, the objective function is the maximization of the ratio $det(S_b)/det(S_w)$. As proven in [41], if $S_w$ is a non-singular matrix, then the ratio is maximized when the column vectors of the projection matrix are the eigenvectors of $S_w^{-1}S_b$. Nevertheless, the non-singularity of the $S_w$ matrix requires at least *N+C* samples, which in many realistic applications is not achievable due to the smaller data set (observations) compared to data dimensionality (*N*). So, the original *N*-dimensional space is projected onto an intermediate lower dimensional space using PCA, and then LDA is used [42]. In this context, LDA is used as feature reduction method.

### D. Wavelet Decomposition

Wavelet analysis is used to decompose the original signal into a set of coefficients that describe the signal frequency content at given times. A wavelet transform uses *wavelets* [43], which are scaled and translated copies of a basic wavelet shape called the 'mother wavelet', to transform the input signals. Mother wavelets are functions localized in both time and frequency and have varying amplitudes during a limited time period and very low or zero amplitude outside that time frame. Wavelet transform yields *wavelet coefficients* that represent the signal in both time and frequency domains. Wavelet transform method is classified into two categories: Continuous Wavelet Transform (CWT) and Discrete Wavelet Transform (DWT), the latter including the Packet Wavelet Transform (PWT) extension.

*1) CWT (Continuous Wavelet Transform)*

The CWT of a signal *f(t)* is computed by using the following equation:

$$C_{a,b} = \int_{-\infty}^{+\infty} f(t)\, \Psi_{a,b}(t)\, dt \qquad (7)$$

where *a* and *b* are scale and translation parameters, respectively of the mother wavelet *Ψ(t)*. The parameter *b* shifts the wavelet so that local information around time *t = b* is contained in the transformed function. The parameter *a* controls the window size in which the signal analysis must be performed. In this way, the obtained functional representation can overcomes the missing localization property of the Fourier analysis [43]. The analysis of a signal using the CWT yields a wealth of information.

*2) DWT (Discrete Wavelet Transform)*

In the CWT the signal is analyzed over infinitely many dilations and translations of the mother wavelet, and, clearly, there will be a lot of redundancy. However, it is possible to retain the key features of the transform by considering subsamples of the CWT [44]. This leads to the Discrete Wavelet Transform. In DWT the signals are passed through high and low pass filters in several stages (levels). In the first level, the signal is decomposed into *approximation coefficients* (via filtration, using a low-pass filter) and into *detail coefficients* (by passing it through a high-pass filter). In the subsequent level, the decomposition is done only on the low pass approximation coefficient obtained at the previous level. This process is duplicated until the desired final level is achieved.

*3) PWT (Packet Wavelet Transform)*

The Packet Wavelet Transform (PWT) is an extension of DWT [45]. In PWT, both detail and approximation coefficients are decomposed at each level. For *n* level of decomposition, the PWT produces *$2^n$* different sets of coefficients (*nodes*) as opposed to (*3n+1*) sets for the DWT. So, a more finer study of the signal is achievable. Due to its characteristics, the PWT is generally employed as an efficient method that considers in detail all ranges of spectral sub-band. In this work, we performed a four-level PWT decomposition.

To achieve optimal performance in the wavelet analysis, a suitable mother wavelet function must be employed. In this study different families of wavelets, such as Daubechies, Symlet, Coiflet, were tested to get the best possible results. Nevertheless, most studies of EC signal analysis have concluded that the Daubechies (Db) wavelet family is the most suitable wavelet [46, 47]. So, in this study, due to similar shape to the EC signal the Daubechies orthogonal wavelets, Db5, was employed. In order to obtain an exact reconstruction of the signal, an adequate number of coefficients must be computed. However, the wavelet transform yields a high-dimensional feature vector. Commonly, the classification performance, resulting from using the high dimensionality of a feature vector, is not efficient in terms of both computation cost and classification accuracy [48]. For these reasons, the selection of the optimal dimensionality reduction method for the wavelet analysis is important before the feature vector is applied in the learning parameters of a classifier. Commonly, feature-projections [49], such as PCA or LDA, are the popular ways to reduce the feature vector's dimensions. Another approach that is frequently used for dimensionality reduction is represented by the time/frequency domain extraction method [50]. Many methods have been proposed during the last decades [51]. In this study, in order to represent the time-frequency distribution of the EC signals, the maximum, minimum, and variance of the absolute values of the coefficients in each sub-band were used. In addition, the following statistical features were also employed:

4) *MAV (Mean Absolute Value)*

MAV represents the mean value of the signal calculated on N samples. It is defined by:

$$MAV = \frac{1}{N}\sum_{n=1}^{N}|x_n| \qquad (8)$$

where $x_n$ represents the *n*-th sample of the wavelet coefficients subsets.

5) *SAP (Scale-Averaged Wavelet Power)*

SAP is the weighted sum of the wavelet power spectrum over scales. SAP can be considered as a time series of the average variance in a certain scale. In other words, it is used to examine the fluctuations in power over a range of particular scales. It is defined by:

$$SAP(n) = \frac{1}{M}\sum_{i=1}^{M}|cwt(i,n)|^2 \qquad (9)$$

Where *CWT*s are the wavelet coefficients, *M* represents the scale size and *n* is the time parameter.

6) *Energy and Entropy*

From an energy point of view, the PWT decomposes the signal energy on different time-frequency plain, and the integration of square amplitude of PWT is proportional to the signal power. Entropy is a common method in many fields, especially in signal processing applications, to evaluate and to compare the probability distributions. Shannon entropy is the most commonly used technique.

The energy of a PWT coefficient (*C*) at level *j* and time *k* is given by:

$$Energy_{j,k} = |C_j(k)|^2 \qquad (10)$$

While, the Shannon entropy can be computed using the extracted wavelet packet coefficients, through the following formula:

$$Entropy_j = -\sum_k |C_j(k)|^2 \log|C_j(k)|^2 \qquad (11)$$

*E. CBIR (Content Based Image Retrieval)*

Content Based Image Retrieval (CBIR) is an actively researched area in computer vision whose goal is to find images similar in visual content to a given query from an image dataset [33]. Image analysis can be based on several distinct features such as color [52], texture [53], shape [54] or any other information that can better describe the image. A typical CBIR system extracts features from each image in the dataset and stores them in a database. Then, when similar images are searched using a "query" image, a feature vector is first extracted from this image, and then a distance between the calculated vector and the database image features is computed. Typical distance metrics between the feature vectors include: Canberra distance, Euclidean distance, Manhattan metric, Minkowski metric and others [55]. If the calculated distance is small, the compared images are considered similar. Compared to the traditional methods, which represent image contents by keywords, the CBIR systems are fast and efficient. The main advantage of the CBIR system is that it uses image features rather than images themselves. For these reason, the application areas are numerous and different: remote sensing, geographic information systems, weather forecasting, medical imaging [56] and recently also in image search over the Internet [57, 58].

There are many different implementations of CBIR. Nevertheless, the key to a good retrieval system is to choose the right features that better represent the images while minimizing the computation complexity.

*1) SGD (Shape Geometric Descriptor)*

The SGD aims to measure geometric attributes of an image. There are many different kinds of shape matching methods, and the progress in improving the matching rate has been substantial in recent years. However, these descriptors are categorized into two main groups: *region-based* shape descriptors and *contour-based* shape descriptors [59]. The first method uses all the pixel information within a shape region of an image. Common region-based methods use moment descriptors [60] that include: geometric moments, Legendre moments, Zernike moments and others [61]. Contour-based approaches use only the information related to the boundary of a shape region and do not consider the shape interior content. These include Fourier descriptor, Wavelet descriptors, curvature scale space and shape signatures [62].

Fig. 2 reports some typical geometric parameters for the shape signatures. They include: Area (A), perimeter (P), centroid (G), orientation angle (α), principal inertia axes, width (W), length (L) and surfaces of symmetry ($S_i$) for an equivalent ellipse image region.

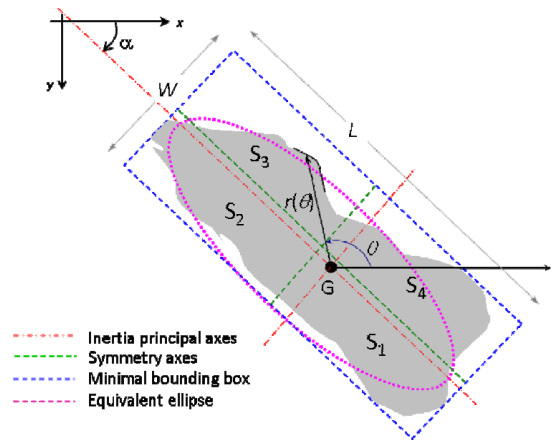

Fig. 2.  Typical geometric parameters.

From these base parameters some advanced parameters (not changing when the original object is submitted to translation, scale changes and rotations) can be derived. They include [63]:

*Compactness*: $C=4\pi A/P^2$. It represents the ratio of the shape area to the area of a circle having the same perimeter.

*Elongation*: $E=L/W$. It is defined by the ratio of the length to the width of the minimal rectangle surrounding the object called also the *minimal bounding box*.

*Rectangularity*: $R=A/(L \times V)$. It quantifies how rectangular a shape is. It is equal to the ratio of the shape area to the area of its minimal bounding box.

*Eccentricity*: It represents the measure of the aspect ratio. It is obtained from the ratio of the minor axis to the major axis in the object equivalent ellipse.

*Convexity*: It is defined as the ratio of perimeters of the convex hull over that of the original contour.

## V. EXPERIMENTAL METHOD AND DATA SETS

We investigated the potential of the soft computing based algorithms when raw data are processed by different feature extraction techniques. In order to provide a proof of concept, we used the resulting procedures to classify the flaws detected by the EC testing.

### A. Ten-fold cross-validation

The classification performance of each classifier is evaluated by using the ten-fold cross-validation method [64], a model validation technique for assessing how the classification results will generalize for an independent data set. Accordingly, all the available data, belonging to the different defects, have been randomly divided into 10 disjoint subsets (folders), each containing approximately the same amount of instances. In each experiment, nine folders have been used as training data, i.e. to set up the classifier, while the remaining folder was used as validation, i.e. to evaluate the classification results. This process was repeated 10 times, for each different choice of validation folder. The 10 results were then averaged to produce a single estimation.

### B. Performance measures

Given a binary classifier and an instance, there are four possible outcomes. If the instance is positive and it is classified as positive, it is counted as a true positive (TP); if it is classified as negative, it is counted as a false negative (FN). If the instance is negative and it is classified as negative, it is counted as a true negative (TN); if it is classified as positive, it is counted as a false positive (FP). Given a classifier and a set of instances (the test set), a two-by-two confusion matrix (also called a contingency table) can be constructed representing the dispositions of the set of instances. This matrix forms the basis for many common metrics. Nevertheless, there is no general consensus on which performance metrics should be used over others [65]. Following, the most common metrics are defined [66]:

- *Accuracy,* that is the portion of correctly classified instances:

$$Accuracy = \frac{TP + TN}{TP + TN + FP + FN} \quad (12)$$

- *Sensitivity* (also called Recall or True Positive Rate - TPR), that measures the portion of actual positives which are correctly identified as such:

$$Sensitivity = \frac{TP}{TP + FN} \quad (13)$$

- *Specificity* (also called True Negative Rate - TNR), that measures the portion of negatives which are correctly identified as such:

$$Specificity = \frac{TN}{TN + FP} \quad (14)$$

- *Precision* (also called positive predictive value), that is a measure of actual positives with respect to all the instances classified as positive:

$$Precision = \frac{TP}{TP + FP} \quad (15)$$

- *F-Measure,* that is the harmonic mean of Precision and Sensitivity. It can be used as a single performance measure:

$$F\ Measure = \frac{2\ Sensitivity * Precision}{Sensitivity + Precision} \quad (16)$$

- *AUC* (Area under ROC curve[1]), that is an estimation of the probability that a classifier will rank a randomly chosen positive instance higher than a randomly chosen negative one.

$$AUC = \frac{Sensitivity + Specificity}{2} \quad (17)$$

- *MCC* (Matthews Correlation Coefficient) that correlates the observed and predicted binary classifications by simultaneously considering true and false positives and negatives. It can assume a value between -1 and +1, where +1 represents a perfect prediction, 0 no better than random prediction and -1 indicates total disagreement between prediction and observation:

$$MCC = \frac{TP\ TN - FP\ FN}{\sqrt{(TP + FP)(TP + FN)(TN + FP)(TN + FN)}} \quad (18)$$

---

[1] *ROC curves* are two-dimensional graphs in which *Sensitivity* is plotted on the *Y* axis, and the complement of *Specificity* (i.e. *1-Specificity*) is plotted on the *X* axis. A ROC graph depicts relative trade-offs between benefits (true positives) and costs (false positives).

*C. Sample data*

Given the intended use on FRA materials, the sample data used in the study refer to a subset of a known database of EC signal samples for aluminum aircraft structures [67]. The overall database is divided in 4 parts.

The first *(part 1)* contains 240 records acquired on an aluminum sample with notches of width 0.3 mm, depth 0.4, 0.7, 1, and 1.5 mm perpendicular, depth 0.4, 0.7, 1, and 1.5 mm with an angle of 30 degrees, 0.7, 1 and 1.5 mm with an angle of 60 degrees and 1.5 mm with an angle of 45 degrees.

The second *(part 2)* refers to 150 records, notches of width 0.2 mm, depth 1, 3 and 5 mm, both perpendicular and 45 degrees orientation of a stainless steel structure.

The third *(part 3)* refers to two-layer aluminum aircraft structure with rivets, two notches below the rivets in the first layer (width 0.2 mm, length 2.5 mm, angle 90 degrees and 30 degrees) and two in the second layer (width 0.2 mm, length 2.5 mm and 5 mm, angle 90 degrees), two defect-free rivets.

The latter *(part 4)* refers to four-layer aluminum structure (layer thickness 2,5 mm) with rivets containing 4 notches (width 0.2 mm, length 2.5 mm, angle 90 degrees) below the rivets in the first, second, third or fourth layer, four defect-free rivets.

In this paper we used two dataset belonging to the *part 1*. The first dataset *(Set 1)* includes only two set of samples acquired on the aluminum structure. The first set refers to the notch perpendicular of width 0.3 mm, depth 1.5 mm. The second refers to the notch oblique of width 0.3 mm, depth 1.5 mm and angle of 60 degrees. The second dataset *(Set 2)* includes the entire *part 1*. It contains twelve types of defects (classes). Each class includes 20 signals.

VI. EXPERIMENTS AND RESULTS

*A. FFT-based Experiments*

We used the MATLAB® environment to perform spectrum analysis of the EC signals. Each signal is composed by 4096 samples, acquired at a sampling frequency of 10 kHz for each of the two acquisition channels.

After performing FFT, the frequency scale was divided in 25 classes equally spaced. For each frequency class we valued the minimum, the maximum, the average and the median of the FFT module. Each frequency class was codified by 4 bits in order to have 16 different levels representing the average value of FFT module in each frequency range. The level ranges were adaptively chosen by considering the dynamic range centered around the median. So each EC signal was coded as a 100 bit feature vector.

*1) Set 1*

For the *Set 1*, as evidenced in the graphical representations of the amplitude spectrum of the positive (notch perpendicular) and negative (notch oblique) instances shown in Fig. 3.a and Fig. 3.b respectively, there is a great separation in the amplitude spectrum between the signals that belong to different classes.

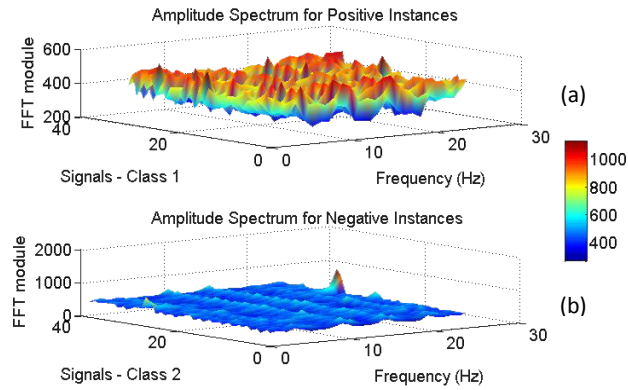

Fig. 3. Amplitude Spectrum of the two datasets belonging to the *Set 1*. The first set refers to the notch perpendicular of width 0.3 mm, depth 1.5 mm (a). The second refers to the notch oblique of width 0.3 mm, depth 1.5 mm and angle of 60 degrees (b).

TABLE I. FFT-BASED EXPERIMENTAL RESULTS -SET 1

| Classifier | Ten-fold cross validation results (means) | | | | | | |
|---|---|---|---|---|---|---|---|
| | *Acc.* | *Sens.* | *Spec.* | *Prec.* | *MCC* | *AUC* | *F-Meas.* |
| J48 | 0.92 | 0.87 | 0.97 | 0.97 | 0.85 | 0.92 | 0.91 |
| Naive Bayes | 0.88 | 0.80 | 0.97 | 0.98 | 0.80 | 0.88 | 0.88 |
| Multilayer Perceptron | 0.98 | 0.97 | 1.00 | 1.00 | 0.97 | 0.98 | 0.98 |
| U-BRAIN | 1.00 | 1.00 | 1.00 | 1.00 | 1.00 | 1.00 | 1.00 |

To set up the soft computing methods described in Section II, we used two set of data, each one of 40 feature vectors, forming the positive and the negative instances required to train the systems.

The performance of the classifiers, C4.5/J48, MLP, NBC and U-BRAIN, evaluated by using the ten-fold cross-validation method (described in Section V), are reported in Table I. The table evidences that the best scores are reached by the U-BRAIN algorithm, reporting the maximum value for all the measures.

The Table II details the ten-fold cross-validation results for U-BRAIN along with the detected formulas (see Appendix I) In the Table, the underscore sign means a literal in negated form [15].

TABLE II. FFT BASED U-BRAIN RESULTS - SET 1

| Test | Ten fold cross validation results | | |
|---|---|---|---|
| | *Rule* | *Training Error* | *Validation Error* |
| 1 | x36 x37 x_46 + x26 x_30 + x_15 x68 x_90 + x8 x_44 + x50 x_87 | 0.00 | 0.00 |
| 2 | x_15 x68 x_71 x83 + x_46 x78 x80 + x_35 x37 x67 + x_11 x14 x37 + x_6 x36 x91 + x_41 | 0.00 | 0.00 |
| 3 | x36 x37 x_46 + x_22 x37 x58 + x28 x_51 x_90 + x_41 x_44 + x7 x_35 x50 | 0.00 | 0.00 |
| 4 | x_15 x_51 x68 + x36 x37 x_46 + x_41 x_63 + x32 x_44 + x5 x_38 | 0.00 | 0.00 |
| 5 | x_15 x68 x_71 x83 + x36 x_46 x66 x100 + x37 x68 x_70 + x32 x_44 + x5 x_96 | 0.00 | 0.00 |
| 6 | x36 x37 x_46 + x_30 x59 x66 x68 + x78 x_92 x100 + x32 x_76 x_90 + x5 x_44 | 0.00 | 0.00 |
| 7 | x_30 x37 x_54 x68 + x17 x36 x_46 x78 + x_30 x_51 x64 x66 + x_44 x50 + x_41 | 0.00 | 0.00 |
| 8 | x14 x36 x66 x_71 + x_41 + x37 x68 x_70 + x37 x_44 + x7 x36 x55 x100 | 0.00 | 0.00 |
| 9 | x_30 x37 x_54 x68 + x36 x_46 x55 + x36 x55 x_80 + x_7 x_11 x_19 + x28 x50 x_60 + x_41 | 0.00 | 0.00 |
| 10 | x_15 x68 x_71 x83 + x_22 x37 x84 + x_18 x_76 x_90 + x36 x_48 x55 + x24 x31 x48 x51 + x_41 | 0.00 | 0.00 |
| Mean | | 0.00 | 0.00 |

*2) Set 2*

The *Set 2* includes 12 defect classes and each class is composed by 20 signals. In each experiment we compared each defect (class) to all the others. In this way we have 12 different set of 20 positive instances and 220 (i.e. 11 x 20) negative ones.

The mean performance of the classifiers, C4.5/J48, MLP, NBC and U-BRAIN, evaluated by using the ten-fold cross-validation method, and averaged over the 12 experiments are reported in Table III. The table highlights a dramatic decrease in performance: there is not a general separation in the amplitude spectrum between the signals that belong to different classes.

TABLE III.  FFT-BASED EXPERIMENTAL RESULTS - SET 2

| Classifier | Ten-fold cross validation results (means) | | | | | | |
|---|---|---|---|---|---|---|---|
| | *Acc.* | *Sens.* | *Spec.* | *Prec.* | *MCC* | *AUC* | *F-Meas.* |
| J48 | 0.85 | 0.24 | 0.90 | 0.23 | 0.14 | 0.57 | 0.34 |
| Naive Bayes | 0.86 | 0.13 | 0.93 | 0.17 | 0.07 | 0.48 | 0.29 |
| Multilayer Perceptron | 0.87 | 0.11 | 0.94 | 0.14 | 0.05 | 0.52 | 0.33 |
| U-BRAIN | 0.85 | 0.22 | 0.91 | 0.20 | 0.12 | 0.56 | 0.33 |

*B. Wavelet-based Experiments*

In this approach, the wavelet coefficients of the EC signals provide a compact representation that shows the energy distribution of the EC signal in time and frequency. These coefficients represent the feature vectors.

In order to reduce the feature vector dimension, we applied the PCA and LDA transformations in a cascade over the set of the wavelet coefficients. Then, statistics over the obtained data were calculated. Accordingly, the feature extraction was accomplished by using both the SAP and MAV values.

*1) CWT-based MAV-SAP classification*

For each raw signal, a continuous Wavelet transformation was performed. Then, we joined MAV and SAP of the wavelet coefficients together to make the feature vector resulting from a single signal. So, with a scale ranging from 1 to 100 and 4096 data samples, we had a feature vector of 4196 elements (4096 for SAP and 100 for MAV) for each raw signal.

The vector dimension, for the *Set 1*, was reduced to 4 elements by the PCA. In fact, we evaluated that the 90% of energy was focused in the 4 highest eigenvalues. In Fig. 4 the first three principal components are depicted for each signal.

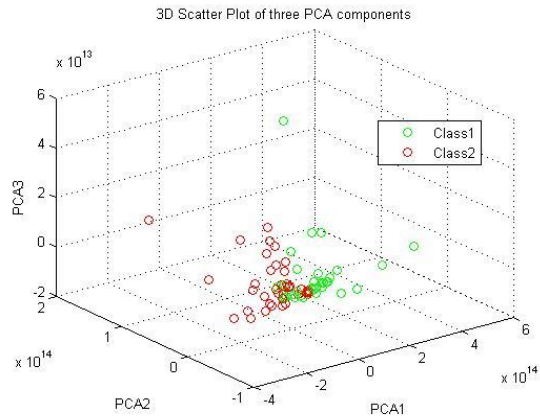

Fig. 4. Scatter graph of the first three PDA-based feature vectors for the 2 classes belonging to the *Set 1*.

Then, a LDA was applied on these vectors, obtaining a single coefficient for each signal. Fig. 5 shows the LDA class separation for the notch perpendicular of width 0.3 mm and the notch oblique of width 0.3 mm and angle of 60 degrees (*Set 1*).

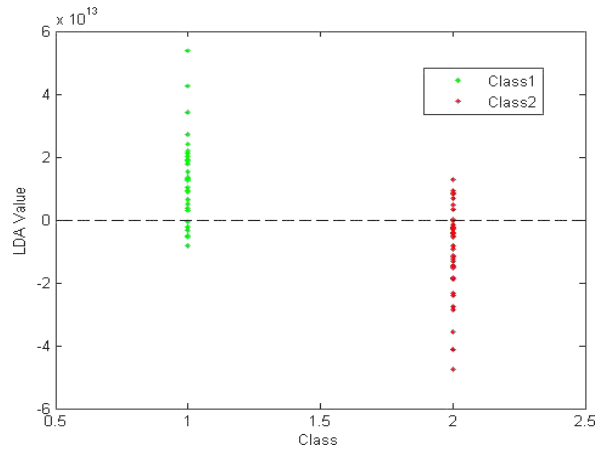

Fig. 5. Plot of the LDA based feature vectors for the two classes of the *Set 1*. Class 1 represents the notch perpendicular of width 0.3 mm. Class 2 represents the notch oblique of width 0.3 mm and angle of 60 degrees. Each dot represents a signal signature.

The one-dimension feature vectors obtained were used as the input to the soft computing classifiers. The performance results are summarized in Table IV. The results are quite poor, and this is mostly due to the overlapping of the classes, as evidenced in Fig. 5, for LDA values around 0.

TABLE IV. CWT MAV-SAP-BASED EXPERIMENTAL RESULTS – SET1

| Classifier | Ten-fold cross validation results (means) | | | | | | |
|---|---|---|---|---|---|---|---|
| | *Acc.* | *Sens.* | *Spec.* | *Prec.* | *MCC* | *AUC* | *F-Meas.* |
| J48 | 0.72 | 0.53 | 0.88 | 0.77 | 0.44 | 0.70 | 0.63 |
| Naive Bayes | 0.82 | 0.81 | 0.83 | 0.79 | 0.64 | 0.82 | 0.80 |
| Multilayer Perceptron | 0.80 | 0.75 | 0.83 | 0.77 | 0.60 | 0.79 | 0.76 |
| U-BRAIN | 0.69 | 0.63 | 0.73 | 0.68 | 0.39 | 0.68 | 0.65 |

To make a performance comparison between LDA-based classification and PCA-only-based classification, we repeated the experiments by using the 4-dimension feature vectors resulting from PCA application as input to the classifiers. The results were worse than before.

TABLE V. CWT MAV-SAP-BASED EXPERIMENTAL RESULTS – SET 2

| Classifier | Ten-fold cross validation results (means) | | | | | | |
|---|---|---|---|---|---|---|---|
| | *Acc.* | *Sens.* | *Spec.* | *Prec.* | *MCC* | *AUC* | *F-Meas.* |
| J48 | 0.83 | 0.13 | 0.90 | 0.11 | 0.03 | 0.51 | 0.29 |
| Naive Bayes | 0.88 | 0.10 | 0.95 | 0.11 | 0.04 | 0.52 | 0.37 |
| Multilayer Perceptron | 0.85 | 0.11 | 0.92 | 0.09 | 0.02 | 0.52 | 0.28 |
| U-BRAIN | 0.83 | 0.13 | 0.89 | 0.10 | 0.02 | 0.51 | 0.25 |

For the *Set 2* the PCA process led to a 6-dimension feature vector for each signal, while the LDA reduced the dimension to 3, corresponding to the 92.3% of the overall energy. The results, depicted in Table V, are also unsatisfactory. In particular, the correlation scores are very low.

*2) CWT-based PCA-LDA classification*

In this experiment the scale-samples matrix of the wavelet coefficients for each signal was arranged to form a unique column vector of 409600 (100*4096) elements.

These column vectors were joined together to make a 409600x72 matrix for the *Set 1* and 409600x434 matrix for the *Set 2* respectively; each column representing a single signal.

Then, the PCA process was employed so reducing the column vector dimension to 9 elements, including the 92% of overall signals energy.

Finally, the LDA process led each vector in the *Set 1* to 1 element, and each vector in the *Set 2* to 4 elements.

The classification results, reported in Table VI and Table VII, for the *Set 1* and the *Set 2* respectively, show low correlation coefficients (*MCC, AUC* and *F-Measure*), even though, at least for the *Set 1,* the other parameters could be acceptable.

TABLE VI. CWT PCA-LDA-BASED EXPERIMENTAL RESULTS – SET 1

| Classifier | Ten-fold cross validation results (means) | | | | | | |
|---|---|---|---|---|---|---|---|
| | *Acc.* | *Sens.* | *Spec.* | *Prec.* | *MCC* | *AUC* | *F-Meas.* |
| J48 | 0.71 | 0.38 | 0.98 | 0.92 | 0,45 | 0.68 | 0.53 |
| Naive Bayes | 0.69 | 0.53 | 0.83 | 0.71 | 0.38 | 0.68 | 0.61 |
| Multilayer Perceptron | 0.69 | 0.47 | 0.88 | 0.75 | 0.38 | 0.67 | 0.58 |
| U-BRAIN | 0.51 | 0.55 | 0.48 | 0.50 | 0.30 | 0.52 | 0.52 |

TABLE VII. CWT PCA-LDA-BASED EXPERIMENTAL RESULTS – SET 2

| Classifier | Ten-fold cross validation results (means) | | | | | | |
|---|---|---|---|---|---|---|---|
| | *Acc.* | *Sens.* | *Spec.* | *Prec.* | *MCC* | *AUC* | *F-Meas.* |
| J48 | 0.84 | 0.15 | 0.91 | 0.13 | 0.05 | 0.53 | 0.33 |
| Naive Bayes | 0.89 | 0.13 | 0.95 | 0.17 | 0.09 | 0.54 | 0.41 |
| Multilayer Perceptron | 0.86 | 0.12 | 0.92 | 0.11 | 0.04 | 0.52 | 0.30 |
| U-BRAIN | 0.84 | 0.14 | 0.90 | 0.12 | 0.04 | 0.52 | 0.29 |

*3) DWT-based classification*

Here, we used the DWT for noise reduction by using twelve decomposition levels, so obtaining 1 approximation coefficient (last level) and 12 detail coefficients.

Each coefficient was represented by the couple made up by its MAV and variance. In this way we obtained a 26-dimension vector for each signal.

Even though the vector dimension is not large, it was further reduced by applying the PCA process. The resulting 3-dimension feature vectors represented the 92% of overall signals energy for the *Set 1,* and the 98.3% for the *Set 2,* respectively.

Finally, we used these vectors as input to classifiers. Tables VIII and IX show the performance results.

TABLE VIII. DWT PCA-BASED EXPERIMENTAL RESULTS – SET 1

| Classifier | Ten-fold cross validation results (means) | | | | | | |
|---|---|---|---|---|---|---|---|
| | *Acc.* | *Sens.* | *Spec.* | *Prec.* | *MCC* | *AUC* | *F-Meas.* |
| J48 | 0.81 | 0.75 | 0.85 | 0.80 | 0.60 | 0.80 | 0.77 |
| Naive Bayes | 0.61 | 0.91 | 0.38 | 0.54 | 0.32 | 0.64 | 0.67 |
| Multilayer Perceptron | 0.71 | 0.63 | 0.78 | 0.69 | 0.41 | 0.70 | 0.66 |
| U-BRAIN | 0.66 | 0.56 | 0.76 | 0.69 | 0.37 | 0.67 | 0.62 |

TABLE IX. DWT PCA-BASED EXPERIMENTAL RESULTS – SET 2

| Classifier | Ten-fold cross validation results (means) | | | | | | |
|---|---|---|---|---|---|---|---|
| | *Acc.* | *Sens.* | *Spec.* | *Prec.* | *MCC* | *AUC* | *F-Meas.* |
| J48 | 0.48 | 0.14 | 0.90 | 0.13 | 0.04 | 0.52 | 0.30 |
| Naive Bayes | 0.90 | 0.06 | 0.97 | 0.09 | 0.01 | 0.51 | 0.35 |
| Multilayer Perceptron | 0.86 | 0.12 | 0.93 | 0.13 | 0.05 | 0.52 | 0.31 |
| U-BRAIN | 0.82 | 0.14 | 0.88 | 0.10 | 0.02 | 0.51 | 0.27 |

*4) PWT-based classification*

In this work, we performed a four-level PWT decomposition. The level of PWT was determined through a trial and error methodology [68]. For each of the $2^4$=16 nodes of the last level we valued MAV, variance and entropy so obtaining a vector of 3x16=48 components for each signal. Then PCA was used as dimension reduction method. The resulting feature vector dimension was of 1 element, including 100% of overall signal energy.

Tables X and XI show the performance results for the *Set 1* and the *Set 2,* respectively, by applying classifiers.

TABLE X. PWT PCA-BASED EXPERIMENTAL RESULTS – SET 1

| Classifier | Ten-fold cross validation results (means) | | | | | | |
|---|---|---|---|---|---|---|---|
| | *Acc.* | *Sens.* | *Spec.* | *Prec.* | *MCC* | *AUC* | *F-Meas.* |

| Classifier | Ten-fold cross validation results (means) | | | | | | |
|---|---|---|---|---|---|---|---|
| | *Acc.* | *Sens.* | *Spec.* | *Prec.* | *MCC* | *AUC* | *F-Meas.* |
| J48 | 0.64 | 0.28 | 0.93 | 0.75 | 0.28 | 0.60 | 0.41 |
| Naive Bayes | 0.49 | 0.78 | 0.25 | 0.45 | 0.04 | 0.52 | 0.57 |
| Multilayer Perceptron | 0.46 | 0.38 | 0.53 | 0.39 | -0.10 | 0.45 | 0.38 |
| U-BRAIN | 0.56 | 0.53 | 0.58 | 0.54 | 0.13 | 0.56 | 0.55 |

TABLE XI.  PWT PCA-BASED EXPERIMENTAL RESULTS – SET 2

| Classifier | Ten-fold cross validation results (means) | | | | | | |
|---|---|---|---|---|---|---|---|
| | *Acc.* | *Sens.* | *Spec.* | *Prec.* | *MCC* | *AUC* | *F-Meas.* |
| J48 | 0.85 | 0.13 | 0.92 | 0.14 | 0.05 | 0.52 | 0.31 |
| Naive Bayes | 0.91 | 0.03 | 0.99 | 0.23 | 0.08 | 0.51 | 0.41 |
| Multilayer Perceptron | 0.87 | 0.11 | 0.94 | 0.15 | 0.06 | 0.52 | 0.34 |
| U-BRAIN | 0.82 | 0.14 | 0.89 | 0.10 | 0.03 | 0.51 | 0.27 |

Overall, the Wavelet-based methods do not appear to grasp the invariant characteristics (signatures) of each class with respect to the other defects considered.

*C. CBIR-based Experiments*

In EC signals the presence of damage is characterized by a particular output probe impedance, resulting in a specific shape in the complex plane.

We used the shape of the impedance in the complex plane to identify defects.

In Fig. 6 a typical shape of coil impedance in the complex plane for an aluminum sample with notch perpendicular of width 0.3 mm and depth 1 mm is shown.

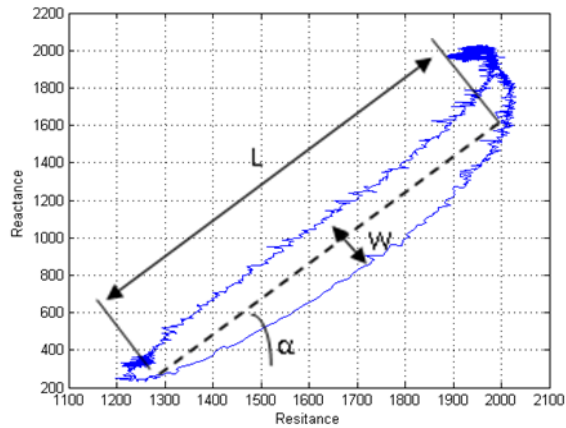

Fig. 6. Typical shape of coil impedance in the complex plane. The dotted line represents the principal inertia axis. (L,W,α) is the feature vector used as the input to classifiers.

As depicted, we intercepted the principal inertia axis and we used the set (L,W,α), composed by the length (L), width (W) and orientation angle (α) of the shape, as feature vector.

*1) Pre-Processing*

To obtain an image suitable for feature extraction, we removed the irrelevant parts from the shapes.

As depicted in Fig. 6 the highest noise was concentrated on the top-right side which represented a high value for both real and imaginary part of the coil impedance. This is also confirmed by the spectrum analysis of each single channel of the samples.

Using the MATLAB® environment, the upper-right side of the image was removed through sorting and cutting procedures acting on the raw data.

Then, by using the Image Processing toolbox, the centroid, the principal inertia axis and then the feature set (L,W,α) were calculated for each record.

As evidenced by the tridimensional scatter graph reported in Fig. 7, the signal representation by means of the proposed feature vectors led to a dramatic separation among classes.

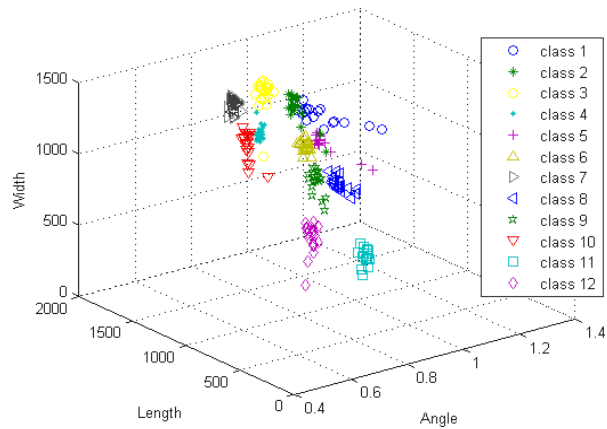

Fig. 7. Tridimensional scatter graph of the proposed feature vectors for 12 classes (*Set 2*). The *Set 1* is represented by class 1 and class 2.

*2) Classification results*

The performance results for *Set 1* and *Set 2*, are reported in Table XII and Table XIII respectively.

TABLE XII. CBIR-BASED EXPERIMENTAL RESULTS – SET 1

| Classifier | Ten-fold cross validation results (means) | | | | | | |
|---|---|---|---|---|---|---|---|
| | *Acc.* | *Sens.* | *Spec.* | *Prec.* | *MCC* | *AUC* | *F-Meas.* |
| J48 | 0.86 | 0.81 | 0.90 | 0.87 | 0.72 | 0.86 | 0.84 |
| Naive Bayes | 0.86 | 0.75 | 0.95 | 0.92 | 0.72 | 0.85 | 0.83 |
| Multilayer Perceptron | 1.00 | 1.00 | 1.00 | 1.00 | 1.00 | 1.00 | 1.00 |
| U-BRAIN | 0.90 | 0.90 | 0.95 | 0.95 | 0.93 | 0.93 | 0.90 |

TABLE XIII. CBIR-BASED EXPERIMENTAL RESULTS – SET 2

| Classifier | Ten-fold cross validation results (means) | | | | | | |
|---|---|---|---|---|---|---|---|
| | *Acc.* | *Sens.* | *Spec.* | *Prec.* | *MCC* | *AUC* | *F-Meas.* |
| J48 | 0.96 | 0.74 | 0.98 | 0.81 | 0.75 | 0.86 | 0.77 |
| Naive Bayes | 0.95 | 0.68 | 0.97 | 0.67 | 0.63 | 0.83 | 0.67 |
| Multilayer Perceptron | 0.98 | 0.85 | 0.99 | 0.89 | 0.84 | 0.92 | 0.87 |
| U-BRAIN | 0.92 | 0.87 | 0.97 | 0.97 | 0.85 | 0.92 | 0.92 |

The experiment results confirmed the perceptible class separation shown in Fig. 7. In particular, the performance parameters related to MLP and U-BRAIN, were very high (near to 1) for both the sets.

*D. Comparison of the Results*

In this subsection, we present a comparison of the adopted technique results on *Set 1* and *Set 2*. Figures 8-21 show, for each adopted performance measure, a summary of the results (*Y-axis*) obtained by varying the feature extraction methods for each machine-learning based algorithm considered (*X-axis*).

  *1) Set 1*

The *Set 1* results are reported in Figures 8-14.

By applying the FFT-based feature extraction method on *Set 1*, the best performance was obtained by the U-BRAIN algorithm.

All the correlations (MCC, AUC, F-Measure), equal to 1 for U-BRAIN, confirmed the excellent ratio between predicted and actually observed classifications. The FFT method appeared to be effective also for MLP, while its outcomes are slightly fewer than U-BRAIN. MLP outperforms J48 and Naïve Bayes.

Wavelet preprocessing showed to be less effective than the FFT. This is probably due to the fact that the statistical coefficients (MAV, SAP, etc.) derived from the discrete wavelet transform tend to reduce the higher frequencies, which could contains useful information.

The PWT based results were found to be the worst for all the classifiers both in terms of performance coefficients and of correlation coefficients. DWT led to acceptable results for J48 and U-BRAIN algorithms.

CWT SAP-MAV-based feature extraction methods was overall effective for all the classifiers.

For the CWT PCA-LDA-based feature extraction method, only specificity was found acceptable, while correlation coefficients were unsuitable.

The CBIR-based classification outperformed the wavelet based techniques and its performance coefficients were found close to the FFT-based ones (11% lower on average). Also in this case the U-BRAIN and the MLP were found to be the most effective algorithms. The U-BRAIN algorithm correlation coefficients were found to be close to the MLP technique. A lower value (16% on average) was found for the J48 and Naïve Bayes algorithms.

Overall, for the *Set 1*, the FFT and CBIR based feature extraction methods appeared as the most effective, and U-BRAIN and MLP were found to be the most adequate classifiers.

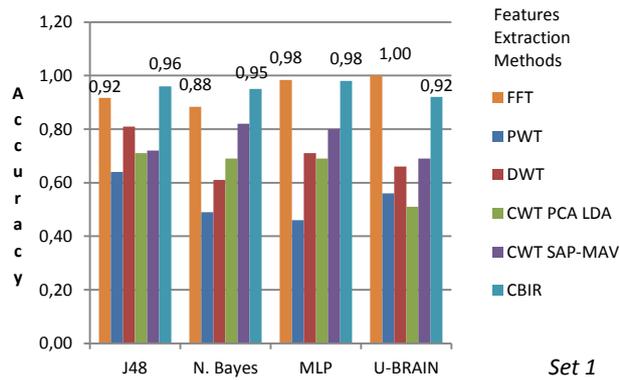

Fig. 8. Accuracy values for different features extraction methods and soft computing based algorithms – *Set 1*.

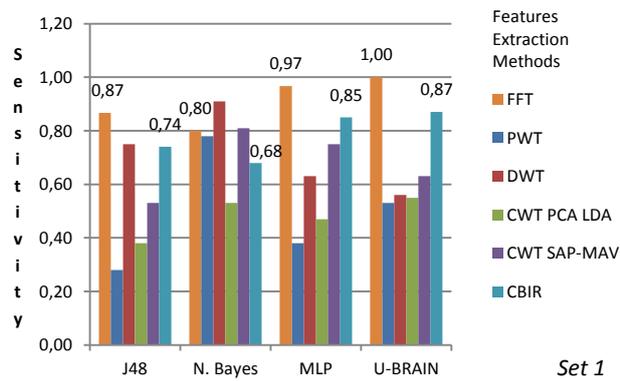

Fig. 9. Sensitivity values for different features extraction methods and soft computing based algorithms – *Set 1*.

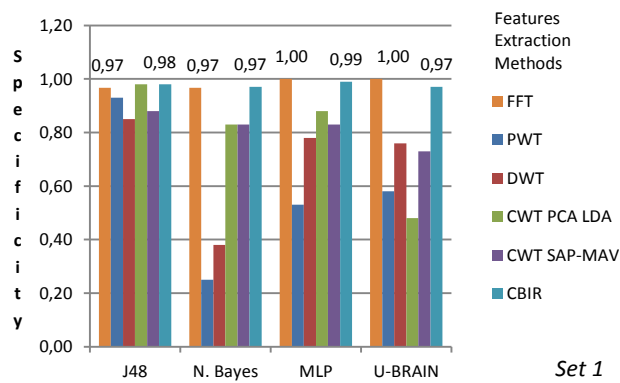

Fig. 10. Specificity values for different features extraction methods and soft computing based algorithms – *Set 1*.

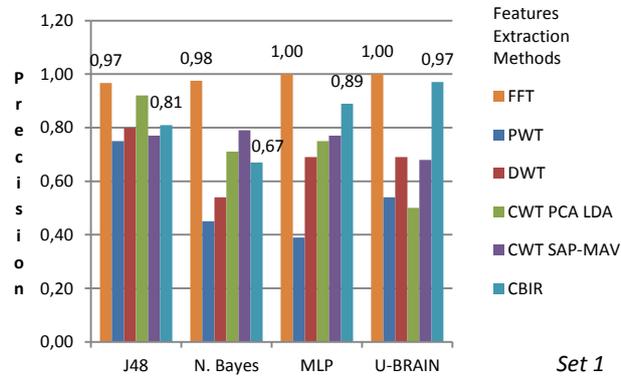

Fig. 11. Precision values for different features extraction methods and soft computing based algorithms – *Set 1*.

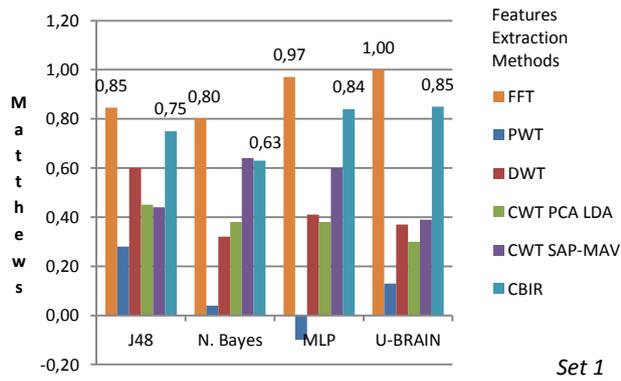

Fig. 12. Matthews correlation coefficients for different features extraction methods and soft computing based algorithms – *Set 1*.

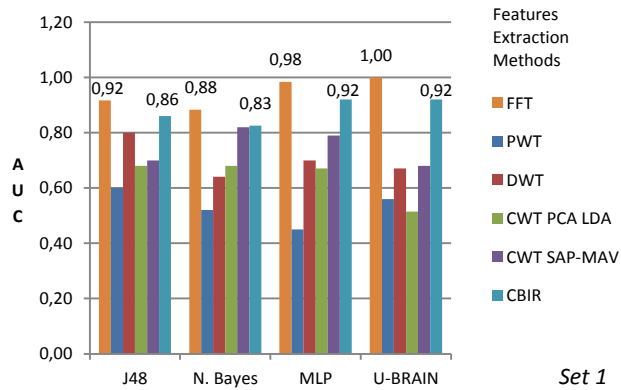

Fig. 13. AUC scores for different features extraction methods and soft computing based algorithms – *Set 1*.

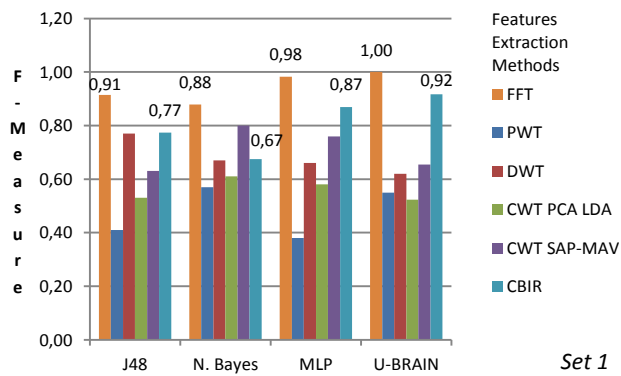

Fig. 14. F-Measure values for different features extraction methods and soft computing based algorithms – *Set 1*.

*2) Set 2*

Figures 15-21 show the results on Set 2.

The Specificity values for the FFT and Wavelet-based methods (Fig. 17) were quite high for all the classifiers. Nevertheless, the low values of the Precision (Fig. 18) and the very low values of the correlation coefficients (Fig. 19-21) evidence the ineffectiveness of the methods.

On the other hand, very high performance coefficients were found for the CBIR method. It outperformed all the other methods for each classifier applied. Also in this case the U-BRAIN and the MLP were found to be the most efficient classifiers.

From the cross comparison between the performance results obtained on the *Set 1* and on the *Set 2* we can conclude that the CBIR is to be considered as the best method for the EC-based defect classification.

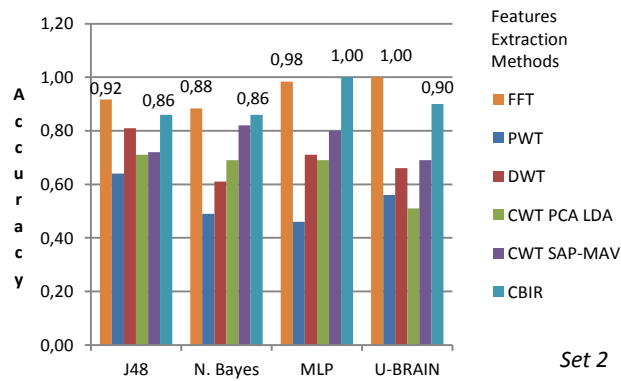

Fig. 15. Accuracy values for different features extraction methods and soft computing based algorithms – *Set 2*.

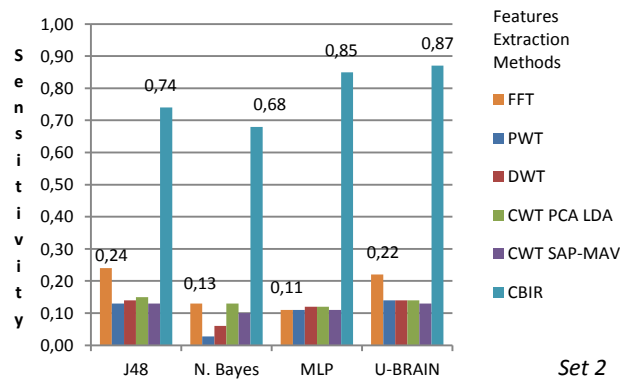

Fig. 16. Sensitivity values for different features extraction methods and soft computing based algorithms – *Set 2*.

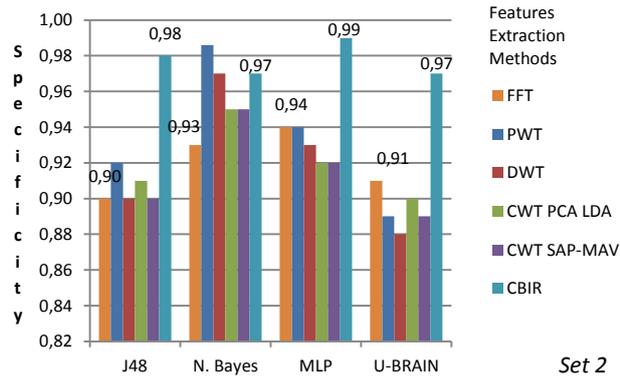

Fig. 17. Specificity values for different features extraction methods and soft computing based algorithms – *Set 2*.

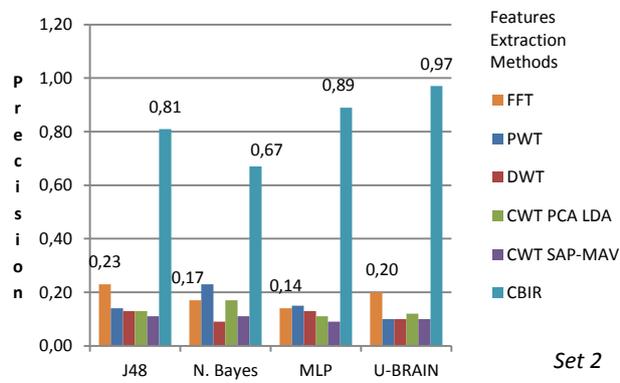

Fig. 18. Precision values for different features extraction methods and soft computing based algorithms – *Set 2*.

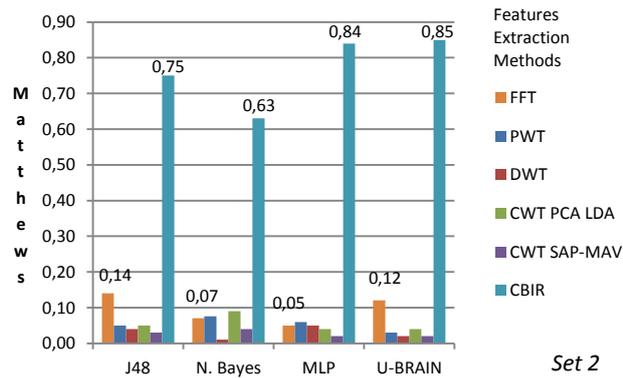

Fig. 19. Matthews correlation coefficients for different features extraction methods and soft computing based algorithms – *Set 2*.

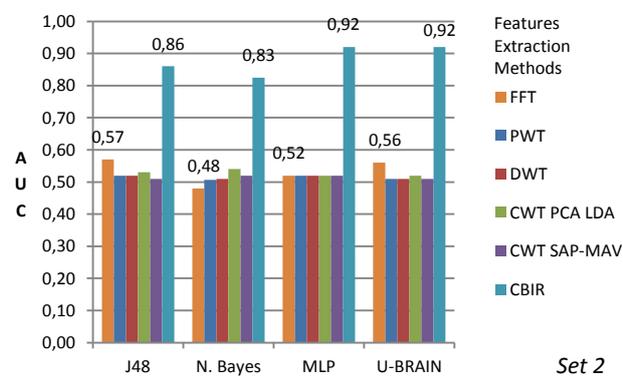

Fig. 20. AUC scores for different features extraction methods and soft computing based algorithms – *Set 2*.

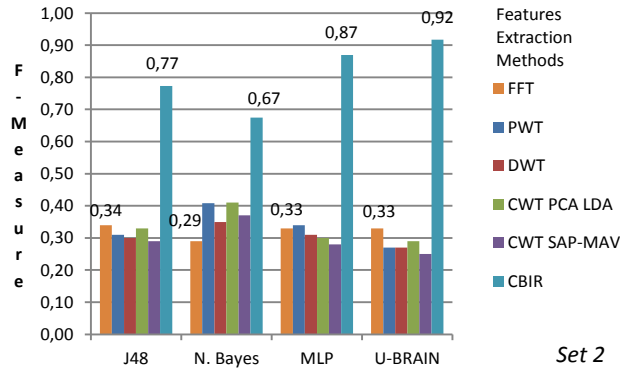

Fig. 21. F-Measure values for different features extraction methods and soft computing based algorithms – *Set 2*.

VII. CONCLUSIONS

In this paper we have investigated several techniques and methods of signal processing and data interpretation to characterize aerospace structure defects. This study has addressed two among the main issues in aerospace structure defects classification: the feature extraction and the classification method.

This has been done by applying different known feature extraction methods (FFT, and Wavelet) and a novel CBIR-based one. The feature vector dimension has been reduced by using PCA and LDA processes. Then some soft computing techniques including the J48 decision trees, the Multilayer Perceptron neural network, the Naive Bayes classifier and the U-BRAIN learning algorithm have been applied, allowing advanced multi-parameter data processing.

The performance of the resulting detection systems have been measured in terms of Accuracy, Sensitivity, Specificity, and Precision. Their effectiveness has been evaluated by the Matthews correlation, the Area Under Curve (AUC), and the F-Measure. Several experiments have been performed on a standard dataset of eddy current signal samples for aircraft structures.

The CBIR approach introduced, using the signal shape as signature, through a feature vector composed by only three geometric parameters, evidenced itself as the most effective. On the other hand, Wavelet and FFT based methods, while largely used in the literature, showed a quite limited behavior with respect to the CBIR method.

The results of this study have evidenced that the key to a successful soft-computing based testing system is to choose the right feature extraction method, representing the defect as accurately and uniquely as possible in a short time.

From a soft computing point of view, U-BRAIN and MLP have been found as the best classifiers. The U-BRAIN algorithm has the further advantage to showing explicitly the rule underlying the process. Compared to other works on the same data [69] the CBIR-ANN and CBIR-U-BRAIN chains have shown better results.

Open problems rest in the validation of the results using larger datasets, even of FRA materials, and in the extension of the results to other NDT techniques as ultrasound and thermography, and this will be matter of a future work.

APPENDIX I – U-BRAIN ALGORITHM

The U-BRAIN is a learning algorithm originally conceived for recognizing splice junctions in human DNA [70, 71]. Splice junctions are points on a DNA sequence at which "superfluous" DNA is removed during the process of protein synthesis in higher organisms.

The general method used in the algorithm is related to the STAR technique of Michalski [72], to the candidate-elimination method introduced by Mitchell [73], and to the work of Haussler [74]. The algorithm has been extended by using fuzzy sets [75], in order to infer a DNF formula that is consistent with a given set of data which may have missing bits.

The conjunctive terms of the formula are computed in an iterative way by identifying, from the given data, a family of sets of conditions that must be satisfied by all the positive instances and violated by all the negative ones; such conditions allow the computation of a set of coefficients (relevances) for each attribute (literal), that form a probability distribution, allowing the selection of the term literals.

Specifically, the algorithm builds Boolean formula of n literals $x_i$ ($i = \{1,...,n\}$) in DNF form, made up of disjunctions of conjunctive terms, starting from a set $T$ of training data.

The data (instances) in $T$ are divided into two classes, named positive and negative, respectively modeled by the $n$-sized vectors $u_i$ with $i = \{1,...,p\}$ and $v_j$ with $j = \{1,...q\}$, representing the issues to be classified. Each element $u_{ik}$ or $v_{jk}$ with $k =$

*{1,...,n}* can assume values belonging to the set *{1,0,1/2}* respectively associated to positive, negative and uncertain instances. The conjunctive terms of the formula are carried-out in an iterative way by two nested loops (see algorithm schema).

*Algorithm schema*

  Require: p>0, q>0, T={$u_1,…,u_p, v_1,…v_q$}
  1. Initialize $f = \emptyset$
  2. **While** there are positive instances $u_i \in T$
      2.1. Uncertainty Reduction
      2.2. Repetition Deletion
      2.3. Initialize term $m = \emptyset$
      2.4. Build $S_{ij}$ sets from T
      2.5. **While** there are elements in $S_{ij}$
          2.5.1. Compute the $R_{ij}$ relevances
          2.5.2. Compute the $R_i$ relevances
          2.5.3. Compute the R relevances
          2.5.4. Choose Literal *x* with *max* relevance R
          2.5.5. Update term: $m = m \cup \{x\}$
          2.5.6. Update $S_{ij}$ sets
      2.6. Add term *m* to *f*: $f = f \cup \{m\}$
      2.7. Update positive instances
      2.8. Update negative instances
      2.9. Check consistency

The inner cycle refers to the selection of the literals of each formula term, while the outer one is devoted to the terms themselves. In order to build a formula consistent with the given data, U-BRAIN compares each given positive instance with each negative one and builds a family of fuzzy sets of conditions that must be satisfied by at least one of the positive instances and violated by all the negative ones formally defined as:

$$S_{ij} = \{x_k | (u_{ik} > v_{ik}) \, or \, (u_{ik} = v_{ik} = \tfrac{1}{2})\} \cup \{\bar{x}_k | (u_{ik} < v_{ik}) \, or \, (u_{ik} = v_{ik} = \tfrac{1}{2})\} \quad (A.1)$$

In other words, the *k*-th literal belongs to the $S_{ij}$ set if the elements in the position *k*, belonging to the *i*-th positive instance $u_{ik}$ and to the *j*-th negative instance $v_{jk}$ are different or both equal to *1/2*. Starting from these sets $S_{ij}$, the algorithm determines for each literal $x_k$ belonging to them a set of coefficients $R_{ij}$, $R_i$ and $R$, called relevances, forming a probability distribution, where:

$$R_{ij}(x_k) = \frac{\chi_{ij}(x_k)}{\#S_{ij}}; \qquad \#S_{ij} = \sum_{m=1}^{2n}\chi_{ij}(x_m)$$

$$R_i(x_k) = \frac{1}{q}\sum_{j=1}^{q} R_{ij}(x_k) \qquad\qquad (A.2)$$

$$R(x_k) = \frac{1}{p}\sum_{i=1}^{p} R_i(x_k)$$

Where $\chi_{ij}$ is the membership function of the set $S_{ij}$ and $\#S_{ij}$ is the fuzzy cardinality of a subset of a set $S_{ij}$. This allows the selection of the literals on a maximum probability greedy criteria (the literal having maximum relevance value is selected). The goal of such greedy selection is simultaneously covering the maximum number of positive instances with the minimum possible number of literals. Each time a literal is chosen, the condition sets $S_{ij}$, and the corresponding probability distribution, are updated by erasing the sets containing the literal itself. The inner cycle is then repeated and the term is completed when there are no more elements in the sets of conditions. Then the new term is added to the formula and, in the outer cycle, the positive instances satisfying the term are erased. Then, the inner cycle starts again on the remaining data. The algorithm ends when there are no more data to treat. The algorithm has two biases: the instance set must be self-consistent, that means that an instance cannot belong to both the classes, and no duplicated instances are allowed. In fact, it may happen that the initial set of training instances contains redundant information. This may be due to repeated instances present from the beginning of the process or resulting from a reduction step, whose task is limiting the presence of missing bits, by recovering them as possible. Such redundancy is automatically removed by keeping each instance just once and deleting all the repetitions, in order to avoid consistency violation that can halt the process.

A. *U-BRAIN Algorithm Complexity*

According to the Landau's symbol [76] to describe the upper bound complexity with big O notation, the overall algorithm time complexity is $\approx O(n^5)$ and the space complexity is in the order of $\approx O(n^3)$ for large *n* (where *n* is the number of variables). So, storing and computing for large data in a computer is space and time consuming.

Of course, such a complexity is only referred to the training phase where the set of classification rules is initially built from the training data (see Fig. 1). Once these rules are available the detection activity is extremely simple and fast and hence can be performed in real time by operating on-line on the live data.

In order to overcome the limitations related to high computational complexity in the training phase, recently an high performance parallel based implementation of U-BRAIN has been realized [77]. Mathematical and programming solutions able to effectively implement the algorithm U-BRAIN on parallel computers have been found; a Dynamic Programming model [78] has been adopted. Finally, in order to reduce the communication costs between different memories and, then, to achieve efficient I/O performance, a mass storage structure has been designed to access its data with a high degree of temporal and spatial locality [79].

Then a parallel implementation of the algorithm has been developed by a Single Program Multiple Data (SPMD) [25] technique together to a Message-Passing Programming paradigm [26]. Overall, the results obtained on standard data sets [80, 81] show that the parallel version is up to 30 times faster than the serial one. Moreover, increasing the problem size, at constant number of processors, the speed-up averagely increases.

ACKNOWLEDGMENTS

This work has been supported in part by Distretto Aerospaziale della Campania (DAC) in the framework of the CERVIA project - PON03PE_00124_1.